\documentclass[10pt,twocolumn,letterpaper]{article}

\usepackage{cvpr}              %
\usepackage{lineno}
\usepackage{colortbl}
\usepackage{booktabs}
\usepackage{multirow}
\usepackage{amsmath}
\usepackage{graphicx}
\usepackage{caption}
\usepackage{lipsum}
\usepackage{float}
\usepackage{utfsym}
\usepackage{diagbox}
\usepackage{float} %
\usepackage{placeins} %
\usepackage{diagbox}
\usepackage{pifont}

\definecolor{cvprblue}{rgb}{0.21,0.49,0.74}
\usepackage[pagebackref,breaklinks,colorlinks,allcolors=cvprblue]{hyperref}
\newcommand{\corr}{\ding{61}}%
\def\paperID{2158} %
\def\confName{CVPR}
\def\confYear{2025}

\title{{Sparse2DGS: Geometry-Prioritized Gaussian Splatting for \\ Surface Reconstruction from Sparse Views}}

\author{Jiang Wu\quad Rui Li\quad Yu Zhu\textsuperscript{\corr}\quad Rong Guo\quad Jinqiu Sun\quad Yanning Zhang\textsuperscript{\corr}\\
Northwestern Polytechnical University 
}

\begin{document}

\twocolumn[{
\renewcommand\twocolumn[1][]{#1}
\maketitle
\begin{center}
    \captionsetup{type=figure}
    \includegraphics[scale=0.47]{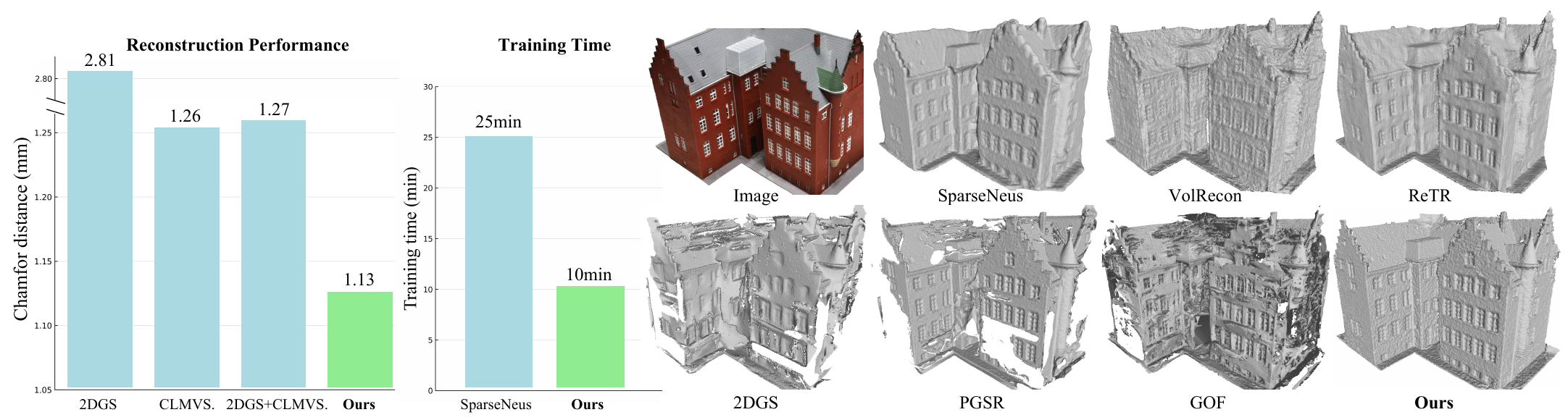}%
    \captionof{figure}{\textbf{Sparse2DGS} boosts the strengths of Gaussian Splatting and Multi-view Stereo in sparse-view surface reconstruction, with notable improvement over 2DGS \cite{huang20242d}, CLMVSNet~\cite{xiong2023cl}, as well as their plain combination (\textit{left}). Meanwhile, Sparse2DGS is ${2} \times$ faster than NeRF-based sparse-view method~\cite{long2022sparseneus} (\textit{middle}) and leads to complete and accurate reconstruction (\textit{right}).
    \label{fig:head}
    }
\end{center}
}]

\begin{abstract}
We present a Gaussian Splatting method for surface reconstruction using sparse input views. Previous methods relying on dense views struggle with extremely sparse Structure-from-Motion points for initialization.
While learning-based Multi-view Stereo (MVS) provides dense 3D points, directly combining it with Gaussian Splatting leads to suboptimal results due to the ill-posed nature of sparse-view geometric optimization. We propose Sparse2DGS, an MVS-initialized Gaussian Splatting pipeline for complete and accurate reconstruction. Our key insight is to incorporate the geometric-prioritized enhancement schemes, allowing for direct and robust geometric learning under ill-posed conditions.
Sparse2DGS outperforms existing methods by notable margins while being ${2}\times $ faster than the NeRF-based fine-tuning approach.
Code is available at \href{https://github.com/Wuuu3511/Sparse2DGS}{https://github.com/Wuuu3511/Sparse2DGS}.

\end{abstract}
\renewcommand{\thefootnote}{}
\footnotetext{\textsuperscript{\corr} indicates corresponding authors.}

\section{Introduction}
\label{sec:intro}
Reconstructing 3D surfaces from multiple views is a fundamental task in computer vision and has been widely used in autonomous driving~\cite{geiger2012we,li2023bevdepth} and virtual reality~\cite{luo2020consistent}. 
While typical methods require abundant input images, reconstructing accurate 3D surfaces from sparse views is particularly challenging and is crucial for scenes that are frequently captured with limited images.
\vspace{-2pt}
\par
NeRF-based surface reconstruction methods~\cite{mildenhall2021nerf} have gained popularity by encoding scene geometry into an occupancy field~\cite{oechsle2021unisurf,li2024know} or a signed distance function (SDF)~\cite{wang2021neus, yariv2021volume}. However, when faced with sparse views, these methods often yield incomplete surfaces with artifacts due to insufficient supervision.
Although sparse-view methods~\cite{long2022sparseneus,ren2023volrecon,liang2024retr,xu2023c2f2neus} use pre-trained feed-forward NeRFs to provide geometric priors, they still exhibit limited generalization to unseen scenes.%

\par
Recently, 3D Gaussian Splatting (3DGS)~\cite{kerbl20233d} has leveraged Structure-from-Motion (SfM) point cloud to construct Gaussian primitives for fast and high-quality novel view synthesis (NVS).
Subsequent works~\cite{huang20242d,yu2024gaussian,chen2024pgsr,chen2023neusg,yu2024gsdf,fan2024trim} aim to achieve more efficient 3D surface reconstruction using 3DGS. Notably, 2D Gaussian Splatting (2DGS) \cite{huang20242d} proposes a 2D variant of Gaussian primitives, with disk-shaped primitives suitable for geometry representation. However, when dealing with sparse views, existing methods~\cite{huang20242d,yu2024gaussian,chen2024pgsr} struggle to produce faithful reconstructions due to initially over-sparse SfM points.

\par
On the other hand, learning-based multi-view stereo (MVS)~\cite{yao2018mvsnet, ding2022transmvsnet, wu2024gomvs,li2023learning} predicts dense depth maps from pixel-wise feature matching, significantly improving geometric completeness even with sparse images. However, MVS depth networks focus on generalizable inference instead of test-time geometry enhancement by design. While combining MVS point cloud with Gaussian Splatting is intuitively effective, their plain combination (see Fig.~\ref{fig:head}, ``2DGS+CLMVS'') yields sub-optimal results due to the ill-posed nature of geometry optimization from sparse views.

\par

In this paper, we propose Sparse2DGS, an MVS-initialized Gaussian Splatting method for complete and accurate sparse-view 3D surface reconstruction. The key idea is to boost the strengths of two paradigms through \textit{geometric-prioritized} enhancement against the ill-posed nature of sparse-view reconstruction.
Starting from MVS initialization, we first incorporate MVS-derived geometric features into the Gaussian splatting process while keeping feature and color values fixed. This provides geometrically expressive supervision and prevents appearance overfitting, thereby improving geometric reconstruction quality.
We further introduce a direct Gaussian primitive regularization, with a reparameterization-based disk sampling strategy that reformulates Gaussian primitives' position, orientation, and scale into point representations. This approach enables a unified optimization of different forms of Gaussian properties within a standard point-based, cross-view consistency loss.
Moreover, instead of densifying Gaussian primitives under ill-posed conditions, we propose to selectively update the MVS-initialized Gaussian primitives with rendered geometric cues, allowing for self-guided geometric corrections.

\par
Sparse2DGS significantly outperforms existing Gaussian Splatting methods in sparse-view surface reconstruction.
Meanwhile, it demonstrates notable efficiency advantages over NeRF-based test-time optimization. 
Our contributions are as follows:
\begin{itemize}[topsep=-0.0cm, itemsep=-0cm]
\item We propose Sparse2DGS, a test-time sparse-view 3D surface reconstruction method that boosts the strengths of Gaussian Splatting and MVS, allowing for complete and accurate reconstruction from sparse images.
\item We focus on {geometric-prioritized} enhancement by proposing geometrically enhanced supervision, direct Gaussian primitive regularization, and selective Gaussian update throughout the Gaussian Splatting pipeline. Sparse2DGS achieves the lowest error ({1.13} CD error) compared to 2DGS (2.81)~\cite{huang20242d}, GOF (2.82)~\cite{yu2024gaussian}, and PGSR (2.08) \cite{chen2024pgsr} on DTU dataset with 3 views. Meanwhile, Sparse2DGS is ${2}\times$ faster than fine-tuning NeRF-based SparseNeus~\cite{long2022sparseneus}.

\end{itemize}

\section{Related Works}
\subsection{3D Reconstruction via Novel View Synthesis}
Recent progresses~\cite{huang2024neusurf,ren2023volrecon,huang20242d,yu2024gaussian} in novel view synthesis (NVS) have enabled accurate 3D surface reconstruction with only image-based supervision.
Neural Radiance Fields~\cite{mildenhall2021nerf} (NeRFs) have become a significant technique for representing 3D scenes.
It has inspired numerous works on directly extracting surfaces by training a parameterized radiance field through differentiable volume rendering.
Neus~\cite{wang2021neus} and VolSDF~\cite{yariv2021volume} integrate surface learning into volume rendering by re-parameterizing the signed distance function as volume density, reconstructing more accurate surfaces.
To reduce geometric deviations and accelerate training, subsequent works~\cite{fu2022geo,sun2022direct,wang2022improved,yu2022monosdf} have introduced improvements in various areas.

Recently, 3D Gaussian Splatting~\cite{kerbl20233d} has shown impressive results in novel view synthesis (NVS), with many recent studies adopting this approach for surface reconstruction.
SuGaR~\cite{guedon2024sugar} aligns 3D Gaussians with scene surfaces by designing a regularization term and extracts an accurate mesh from the Gaussians using Poisson Reconstruction~\cite{kazhdan2006poisson}.
2DGS~\cite{huang20242d} replaces 3D Gaussian ellipsoids with 2D-oriented planar disks, providing view-consistent geometric representations.
RaDe-GS~\cite{zhang2024rade} further designs a rasterized approach to render depth and normal, enhancing geometric accuracy while ensuring rendering efficiency.
GOF~\cite{yu2024gaussian} extracts the level set from trained Gaussian primitives by constructing a Gaussian opacity field.
PGSR~\cite{chen2024pgsr} improves mesh quality by introducing single-view and multi-view geometric constraints.
Additionally, GSDF~\cite{yu2024gsdf} and NeusSG~\cite{chen2023neusg} reconstruct scenes by combining the advantages of 3D Gaussian Splatting and SDF networks.
Despite these methods' success, the extracted geometry tends to be noisy and incomplete when the input number of training viewpoints is limited.

\subsection{Learning-based Multi-view Stereo}
Deep learning-based Multi-view Stereo (MVS) estimates the reference image's depth from muti-view posed images.
MVSNet~\cite{yao2018mvsnet} first introduces the differentiable homography warping to build the cost volume with deep image features and then employs a 3D CNN for cost regularization.
Recent follow-up works have introduced improvements in various aspects, including multi-view feature extraction~\cite{ding2022transmvsnet,cao2022mvsformer}, cost aggregation~\cite{xu2022learning, zhang2023vis,xu2024sdge,xu2022multi}, and cost volume regularization~\cite{wu2024gomvs}.
However, these methods require ground-truth depth maps for training, which are challenging to obtain in practice.
Unsupervised MVS~\cite{xiong2023cl,khot2019learning,chang2022rc} can be applied in more general, real-world scenarios.
MVS methods can infer dense depth maps for each input view and therefore, we select the unsupervised CLMVSNet~\cite{xiong2023cl} to initialize the Gaussian point cloud.

\begin{figure*}[t]
\begin{center}
\includegraphics[width=1\linewidth]{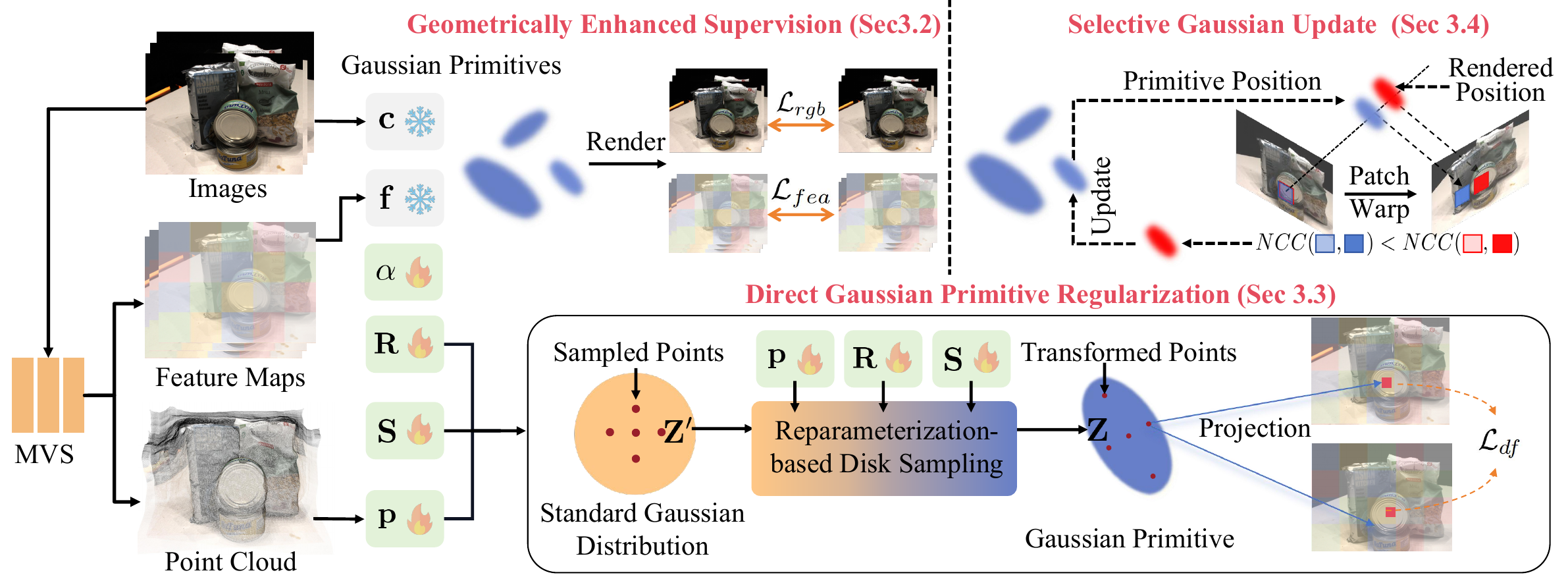}
\end{center}
\vspace{-10pt}
\caption{\textbf{Overview.} 
Given sparse posed images, we first leverage the MVS points to initialize the Gaussian position $\mathbf{p}$ (Sec. \ref{2D Gaussian Splatting with MVS initialization}).
We then leverage the MVS-derived feature for Gaussian Splatting, with fixed feature $\mathbf{f}$ and color $\mathbf{c}$ values to conduct geometrically expressive supervision and avoid appearance overfitting (Sec. \ref{Geometrically Enhanced Supervision}). 
We then optimize Gaussian primitive properties by reformulating orientation $\mathbf{R}$, scale $\mathbf{S}$, and position $\mathbf{p}$ into sampled points using the reparameterization-based disk sampling, ensuring simultaneously optimizing different forms of Gaussian properties with point-based loss (Sec. \ref{Direct Gaussian Primitive Regularization}). 
Finally, we introduce a Selective Gaussian Update strategy that leverages rendered geometry as an alternative to update Gaussian primitive positions, which further improves reconstruction quality (Sec. \ref{Selective Gaussian Update}).}
\vspace{-10pt}
\label{fig:arc}
\end{figure*}

\subsection{Sparse View Surface Reconstruction}
Recent works have explored 3D surface reconstruction from sparse training views.
SparseNeuS~\cite{long2022sparseneus} achieves reasonable mesh reconstruction results by constructing a cascaded volume to learn a generalizable radiance field, followed by per-scene fine-tuning.
NeuSurf~\cite{huang2024neusurf} learns dense on-surface priors from sparse points to guide geometric learning.
SparseCraft~\cite{younes2024sparsecraft} leverages stereopsis cues for geometric regularization based on Taylor expansion.
Moreover, VolRecon~\cite{ren2023volrecon} and ReTR~\cite{liang2024retr} enhance voxel features by designing specialized transformer architectures.
C2F2Neus~\cite{xu2023c2f2neus} constructs per-view cost frustum for finer geometry estimation.
Although these feedforward networks can directly infer scene geometry, their limited generalization to unseen scenes restricts their applicability in real-world scenarios.

\vspace{-3pt}
\section{Methodology}
Given sparse view posed images $\{\mathbf{I}_{i} \}{_{i=1}^{N}}$, where $\mathbf{I}_{i}\in \mathbb{R}^{H\times W \times 3}$ and $N$ denotes the view number, Sparse2DGS aims to reconstruct accurate 3D scene surfaces using Gaussian Splatting.
As shown in Fig.~\ref{fig:arc}, we first initialize the Gaussian primitives with the MVS~\cite{xiong2023cl} reconstructed point cloud. 
Then, we sample MVS geometric feature maps into feature splatting with its value and RGB color fixed for supervision. For each Gaussian primitive, we regularize its geometric properties (position, orientation, scale) by a novel reparameterization-based 2D disk sampling strategy that formulates these properties as point-based supervision.
Finally, we propose a selective Gaussian update strategy to further enhance Gaussian primitives using rendered geometric cues.
\label{sec:methodology}
\subsection{2D Gaussian Splatting with MVS initialization}
\label{2D Gaussian Splatting with MVS initialization}
2D Gaussian Splatting~\cite{huang20242d} (2DGS) represents Gaussian primitives as planar disks suitable for representing and optimizing 3D surfaces. However, it relies on sparse SfM~\cite{schonberger2016structure} 3D points for initialization, which exhibits limited reconstruction quality under sparse-view inputs. 
To address this, we initialize 2DGS with an MVS point cloud to provide dense geometric representations.

\noindent\textbf{2D Gaussian Splatting}. Each 2D Gaussian primitive is characterized by its central position $\mathbf{p}$, two principal tangential vectors  $\mathbf{t_u}$  and  $\mathbf{t_v}$, and a scaling vector  $(s_u, s_v)$.
The rotation matrix of a 2D Gaussian can be represented as  $\mathbf{R} = (\mathbf{t_u}, \mathbf{t_v}, \mathbf{t_z})$, where  $\mathbf{t_z} $ is defined by two orthogonal tangent vectors $\mathbf{t_z} = \mathbf{t_u} \times \mathbf{t_v}$ and the scale matrix is $\mathbf{S} = \text{diag}(s_u, s_v, 0) $.
A 2D Gaussian is then defined in a local tangent plane in world space, which is parameterized:
\begin{equation}
\label{2dgs_parameterized}
P(u, v)
=
\mathbf{p}+{s_u}{\mathbf{t}_u}u+
{s_v}{\mathbf{t}_v}v.
\end{equation}
For the point $\mathbf{u} = (u, v)$ in $uv$ space, its 2D Gaussian value can then be evaluated by standard Gaussian
\begin{equation}
\label{cdf}
\mathcal{G}(\mathbf{u})
=
\text{exp}\frac{-(u^2 + v^2)}{2}.
\end{equation}

\noindent\textbf{Initializing 2DGS with MVS depth}.
MVS methods estimate dense depth maps through cross-view matching. In particular, unsupervised MVS~\cite{xiong2023cl} reconstructs complete scenes without ground truth for training, offering greater flexibility.
Therefore, we use the point cloud reconstructed by the current unsupervised method CLMVSNet~\cite{xiong2023cl} to initialize our Gaussian point cloud. Specifically, given sparse views' images $\{\mathbf{I}_{i} \}{_{i=1}^{N}}$, we use the MVS network for depth maps $\{\mathbf{D}_{i} \}{_{i=1}^{N}}$, then back-project these depth maps and fuse into a point cloud $\{\mathbf{p}_{ij} \mid 1 \leq i \leq N,\; 1 \leq j \leq HW \}$ as initialized positions for Gaussian primitives, where $i$ denotes image index and $j$ is the pixel index within each image. 

\subsection{Geometrically Enhanced Supervision}
\label{Geometrically Enhanced Supervision}
3D/2D Gaussian Splatting~\cite{kerbl20233d,huang20242d} couples color and geometry optimization through image-based loss. However, this leads to severe degradation under sparse views as color will overfit sparse views, leading to a compromised geometry. Meanwhile, the challenges in image areas (\eg, textureless regions) will be amplified with sparse views. 
To address this issue, we propose geometrically enhanced supervision by incorporating off-the-shelf expressive MVS features for feature splatting and supervision, with the color and feature fixed to avoid overfitting.
\par

In addition to providing dense point clouds from input images $\{\mathbf{I}_{i} \}{_{i=1}^{N}}$, the MVS network~\cite{xiong2023cl} offers off-the-shelf feature maps with geometric expressiveness. 
We leverage MVS features to enhance geometric supervision by incorporating additional feature attribute $\textbf{f}_{ij}$ into Gaussian primitives.
Specifically, for each input image $\mathbf{I}_{i}$, we extract its feature map $\textbf{F}_{i}$ from the Feature Pyramid Network (FPN) of CLMVSNet~\cite{xiong2023cl}. 
Since each 3D point corresponds to an image pixel due to MVS initialization, we sample both RGB color and MVS features into each Gaussian primitive.
\begin{equation}
    \mathbf{f}_{ij} = \textbf{F}_{i}(\pi(\textbf{p}_{ij})), \mathbf{c}_{ij} = \textbf{I}_{i}(\pi(\textbf{p}_{ij})),
\end{equation}
where $\pi(\cdot)$ denotes the projection from 3D space onto 2D image plane. We further render the feature into an image plane using the tile-based renderer together with RGB rendering. We show the rendering process given $M$ Gaussian primitives along a ray, with indices $i,j$ omitted for brevity  

\begin{equation}
\label{feature spalting}
\{ \hat{\textbf{c}},  \hat{\textbf{f}} \} 
= \sum_{m=1}^M \left\{ \textbf{c}_m,\textbf{f}_m \right\} \cdot \alpha_m \prod_{o=1}^{m-1}
{(1 - \alpha_m)},
\end{equation}
where $\alpha_m$ is the opacity of the Gaussian. With rendered feature maps that have enhanced geometric expressiveness for the challenging sparse views, we further supervise the rendered features using cosine similarity loss by

\begin{equation}
    \mathcal{L}_{fea} = 1 - \frac{\textbf{f}_{ij} \cdot \hat{\textbf{f}}_{ij}}{\|\textbf{f}_{ij}\| \|\hat{\textbf{f}}_{ij}\|},
\label{fealoss}
\end{equation}

Note that the simultaneous learning of appearance (\ie, color $\textbf{c}_{ij}$ and feature $\textbf{f}_{ij})$ and geometry lead to geometric degradation under sparse views (see Tab.~\ref{table:color}). To address this issue, both $\textbf{f}_{ij}$ and $\textbf{c}_{ij}$ are \textit{fixed} during the optimization to enforce the learning specifically for geometry. 
\par
Our method essentially differentiates previous feature-based methods~\cite{zhou2024feature, qiu2024feature} in its specific strategy for geometric enhancement. In our method, the expressive MVS features are incorporated, and features of each Gaussian are fixed, both for geometry enhancement with sparse views. 
Previous methods leverage splatted features only for segmentation or editing~\cite{zhou2024feature, qiu2024feature} with dense views, while their feature losses do not optimize the scene geometry.

\subsection{Direct Gaussian Primitive Regularization}
\label{Direct Gaussian Primitive Regularization}
\begin{figure}[t]
\begin{center}
\includegraphics[scale=0.55]{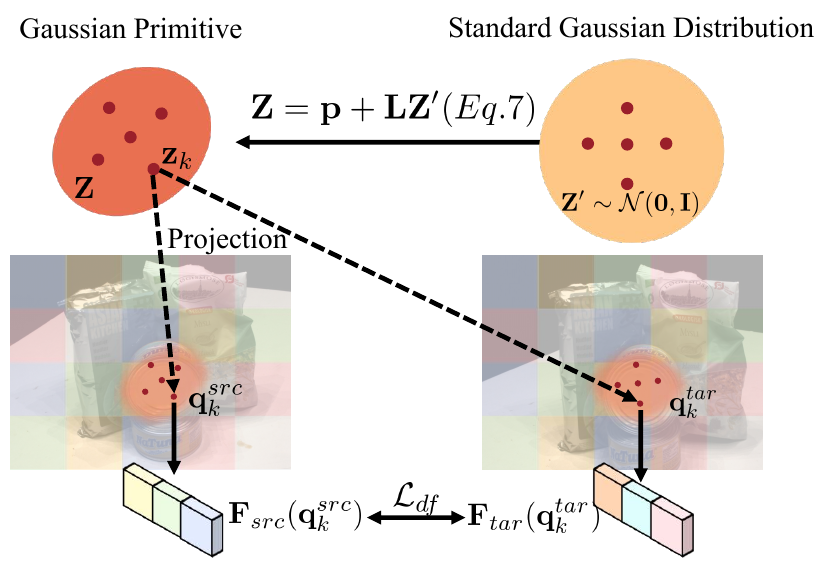}
\vspace{-10pt}
\end{center}
   \caption{\textbf{Direct Gaussian Primitive Regularization.} We sample point set $\mathbf{Z}' \sim \mathcal{N}(\mathbf{0}, \mathbf{I})$ from the standard Gaussian distribution and leverage the reparameterization technique to transform $\mathbf{Z}'$ to the point set $\mathbf{Z}$ on each Gaussian primitive. 
   This process allows for representing Gaussian orientation $\mathbf{R}$ and scale $\mathbf{S}$ properties in the form of points easier for geometric supervision. The transformed points $\mathbf{Z}$ are then projected to the source and target views, supervised by cross-view feature consistency loss. 
   }
   \vspace{-10pt}
\label{fig:diskproject}
\end{figure}

The disk-shaped 2D Gaussian primitives encode latent scene geometry including local shape orientation $\mathbf{R}$, scale $\mathbf{S}$, and position $\mathbf{p}$, which are essential for accurate surface reconstruction. These geometric properties are typically learned from surrogate image render loss supported by abundant multi-view supervision. When facing sparse-view inputs, the primitives tend to degrade due to insufficient geometric constraints.
\par
To address this issue, we incorporate \textit{direct} geometric regularization for 2D Gaussian primitives through multi-view consistency guidance. 
As multi-view consistency approaches only consider point position, we incorporate broad Gaussian primitive properties (\ie, orientation, scale, and position) by correlating their relationship with point position, through a novel reparameterization-based disk sampling strategy. 
This regularization aligns 2D Gaussian primitives more closely with the latent scene surfaces, leading to improved geometric learning.
In the following illustration, we omit the Gaussian primitive indices $i, j$ for brevity. 
\par
\noindent\textbf{Regularizing points with cross-view consistency}. Given 3D point sampling positions $\mathbf{Z}=\{\mathbf{z}_{k}\}_{k=1}^{K}$ on each Gaussian, we project them to source and target views to obtain the corresponding 2D positions $\mathbf{q}_{k}^{src}$ and $\mathbf{q}_{k}^{tar}$, as well as queried MVS features  $\mathbf{F}_{src}(\mathbf{q}_{k}^{src})$, $\mathbf{F}_{tar}(\mathbf{q}_{k}^{tar})$.
We can then optimize the 3D positions by cross-view feature consistency loss~\cite{zhang2021learning} $\mathcal{L}_{df}$:
\begin{equation}
\mathcal{L}_{df} = \frac{1}{K}\sum_{k=1}^{K}(1 - \frac{\textbf{F}_{src}(\textbf{q}^{src}_k) \cdot \textbf{F}_{tar}(\textbf{q}^{tar}_k)}{\|\textbf{F}_{src}(\textbf{q}^{src}_k)\| \|\textbf{F}_{tar}(\textbf{q}^{tar}_k)\|}),
\label{fealoss}
\end{equation}
where $K$ denotes the number of sampled points on each disk-shaped Gaussian primitive.
\par
\noindent\textbf{Reparameterization-based disk sampling.}
The above operation focuses on position regularization only, and thus, it cannot propagate gradients to other Gaussian primitive properties such as orientation $\mathbf{R}$ and scale $\mathbf{S}$. We propose a reparameterization-based sampling strategy to solve this problem.  
As shown in Fig. \ref{fig:diskproject}, for each Gaussian primitive, we first sample points $\mathbf{Z}'$ from the standard Gaussian distribution $\mathcal{N}(\mathbf{0}, \mathbf{I})$ and then transform the distribution to the target Gaussian primitive through its position $\mathbf{p}$, orientation $\mathbf{R}$ and scale $\mathbf{S}$
\begin{equation}
\mathbf{Z} = \mathbf{p} + \mathbf{L} \mathbf{Z'},
\end{equation}
\begin{equation}
\mathbf{L} = \mathbf{R}\mathbf{S},
\end{equation}
where $\mathbf{Z'} \sim \mathcal{N}(\mathbf{0}, \mathbf{I})$.
This bridges the gap between Gaussian properties (orientation and scale) and point representation, ensuring regularizing Gaussian geometric properties merely through the surrogate point-based loss.
Additionally, since the last scale value of the 2D Gaussian primitive~\cite{huang20242d} is zero, resulting in difficulties in optimizing the normal vector $\mathbf{R}[\ldots, 2] $ in the last dimension,
we apply additional supervision using the rendered normal $\mathbf{n}$:
\begin{equation}
\mathcal{L}_{dn}=1 - \mathbf{R}[\ldots, 2]^\top \mathbf{n},
\label{nloss}
\end{equation}
The final primitive regularization loss can be defined as:
\begin{equation}
\mathcal{L}_{dr}=\mathcal{L}_{df} + \mathcal{L}_{dn}.
\label{nloss}
\end{equation}

\subsection{Selective Gaussian Update}
\label{Selective Gaussian Update}
Gaussian Splatting leverages adaptive density control to densify the Gaussian primitives. However, with sparse views, the densified Gaussian primitives are prone to degrade without sufficient multi-view supervision. 
To address this issue, as shown in Fig. \ref{fig:arc}, instead of continually densifying the MVS-initialized dense Gaussian primitives with limited supervision, we propose to selectively update them guided by the alternative rendered geometric cues, \eg, depth and normal.

\noindent{\textbf{Evaluating Gaussian primitives via patch warping.}} 
Given any Gaussian primitive with its position $\mathbf{p}$ and orientation $\mathbf{R}$, we can warp an image patch from one source view to a target view using its position $\mathbf{p}$ and orientation $\mathbf{R}$ together with the pose, following homography warping techniques~\cite{chang2022rc}. Then, we can evaluate the geometric quality of the Gaussian primitive via the patch-based Normalized Cross-Correlation (NCC)~\cite{yoo2009fast} metric.

\begin{equation}
\text{NCC}(x,y) = \frac{\sum_{i=l}^{n} (x_l - \bar{x})(y_l - \bar{y})}{\sqrt{\sum_{l=1}^{n} (x_l - \bar{x})^2} \sqrt{\sum_{l=1}^{n} (y_l - \bar{y})^2}},
\end{equation}
where $x_l$ and $y_l$ represent the pixels in the patches $x$ and $y$ of the source and target images, respectively.

\noindent{\textbf{Update Gaussian primitive with rendered cues.}} 
In addition to using the Gaussian primitive parameter $\mathbf{p}$ and orientation $\mathbf{R}$ for patch warping, we can leverage the rendered pixel-wise depth and normals from tiled Gaussian primitives as an alternative for patch warping. We compute patch-wise NCC over two warping strategies, yielding NCC$_{G}$ for Gaussian primitive and NCC$_{R}$ for rendered cues.
\par
If NCC$_{R}$ is better than NCC$_{G}$, we reproject the rendered depth into a 3D point and update the position of the corresponding Gaussian primitive.
We update the position of Gaussian primitives every 100 training steps, after which all Gaussian parameters continue to be optimized by gradients.
In this way, we improve the initial Gaussian primitive with geometric cues from rendered results, further enhancing the reconstruction quality.

\subsection{Training loss}
During training, in addition to using the disk regularization loss $\mathcal{L}_{dr}$ and MVS feature splatting loss $\mathcal{L}_{fea}$, we retain the original 2DGS \cite{huang20242d} image reconstruction loss $\mathcal{L}_{rgb}$, depth-normal consistency loss $\mathcal{L}_{n}$, and depth distortion loss $\mathcal{L}_{d}$. The final loss function is defined as:
\begin{equation}
\mathcal{L} = \mathcal{L}_{rgb} + \lambda_1  \mathcal{L}_{d} + \lambda_2  \mathcal{L}_{n},
+ \lambda_3 \mathcal{L}_{dr} + \lambda_4 \mathcal{L}_{fea},
\end{equation}
where $\lambda_1, \lambda_2, \lambda_3, \lambda_4$ denotes the hyperparameters for the weights of each loss term.

\begin{table*}[t]
\centering
\renewcommand{\arraystretch}{1}
 \setlength{\tabcolsep}{8pt}
 \scalebox{0.78}{
\begin{tabular}{lcccccccccccccccc}
\toprule
Scan        & 24   & 37   & 40   & 55   & 63   & 65   & 69   & 83   & 97   & 105  & 106  & 110  & 114  & 118  & 122  & Mean \\ \midrule

Colmap~\cite{schonberger2016structure}      &0.90 &2.89 &1.63 &1.08 &2.18 &1.94 &1.61 &1.30 &2.34 &1.28 &1.10 &1.42 &0.76 &1.17 &1.14 &1.52 \\
CLMVSNet~\cite{xiong2023cl}    &1.17&2.52	&1.68	&0.89	&1.51	&1.82	&0.99	&1.29	&1.52	&0.78	&0.88	&1.10	&0.64	&1.04	&1.12 &1.26 \\
														 
Neus~\cite{wang2021neus}        &4.57 &4.49 &3.97 &4.32 &4.63 &1.95 &4.68 &3.83 &4.15 &2.50 &1.52 &6.47 &1.26 &5.57 &6.11 &4.00 \\
Volsdf~\cite{yariv2021volume}      &4.03 &4.21 &6.12 &0.91 &8.24 &1.73 &2.74 &1.82 &5.14 &3.09 &2.08 &4.81 &0.60 &3.51 &2.18 &3.41 \\
VolRecon~\cite{ren2023volrecon} &1.20	&2.59	&1.56	&1.08	&1.43	&1.92	&1.11	&1.48	&1.42	&1.05	&1.19	&1.38	&0.74	&1.23	&1.27 &1.38	  \\
SparseNeus~\cite{long2022sparseneus}  &1.29	&2.27	&1.57	&0.88	&1.61	&1.86	&1.06	&1.27	&1.42	&1.07	&0.99	&0.87	&0.54	&1.15	&1.18	&1.27 \\
Retr~\cite{liang2024retr}        & 1.05 & 2.31 & 1.44 & 0.98 & 1.18 & \textbf{1.52} & 0.88 & 1.35 & 1.30 & 0.87 & 1.07 & 0.77 & 0.59 & 1.05 & 1.12 & 1.17 \\
C2F2Neus~\cite{xu2023c2f2neus}    & 1.12 & 2.42 & \textbf{1.40} & 0.75 & 1.41 & 1.77 & 0.85 & 1.16 & 1.26 & 0.76 & 0.91 & 0.60 & 0.46 & 0.88 & 0.92 & 1.11 \\ 
Sparsecraft~\cite{younes2024sparsecraft}    & 1.17 &\textbf{1.74} & 1.80 & \textbf{0.70} & 1.19 & 1.53 & 0.83 & \textbf{1.05} & 1.42 & 0.78 &0.80 & 0.56 & 0.44 & 0.77 & \textbf{0.84} & 1.04 \\ 

NeuSurf~\cite{huang2024neusurf}     &\textbf{0.78} &2.35 &1.55 &0.75 &\textbf{1.04} &1.68 &\textbf{0.60} &1.14 &\textbf{0.98} &\textbf{0.70} &\textbf{0.74} &\textbf{0.49} &\textbf{0.39} &\textbf{0.75} &0.86 &\textbf{0.99}      \\ \midrule
2DGS~\cite{huang20242d}        &3.14	&3.79	&2.31	&1.51	&5.34	&2.22	&1.85	&2.79	&3.62	&1.63	&3.05	&4.11	&1.05	&3.35	&2.44	&2.81      \\
GOF~\cite{yu2024gaussian}         &3.27	&5.05	&2.77	&2.04	&3.84	&2.61	&2.23	&2.65	&3.30	&1.65	&2.58	&4.44	&1.16	&2.29	&2.39	&2.82     \\
PGSR~\cite{chen2024pgsr}        &3.36	&3.28	&2.75	&1.27	&5.15	&1.84	&0.88	&1.79	&3.49	&1.19	&1.86	&1.11	&0.61	&1.09	&1.52	&2.08     \\ 
\textbf{Ours}        &\textbf{1.05}	&\textbf{2.35}	&\textbf{1.38}	&\textbf{0.83}	&\textbf{1.37}	&\textbf{1.45}	&\textbf{0.84}	&\textbf{1.16}	&\textbf{1.43}	&\textbf{0.74}	&\textbf{0.85}	&\textbf{0.84}	&\textbf{0.57}	&\textbf{0.95}	&\textbf{1.01}   &\textbf{1.13}  \\ \bottomrule
\end{tabular}}
\vspace{-5pt}
\caption{\textbf{Quantitative results on DTU dataset.} Our method achieves the best performance among Gaussian Splatting surface reconstruction approaches and demonstrates competitive performance to previous state-of-the-art NeRF-based methods.}
\vspace{-10pt}
\label{table:dtu_cd}
\end{table*}

\begin{figure*}[t]
\begin{center}
\includegraphics[scale=0.38]
{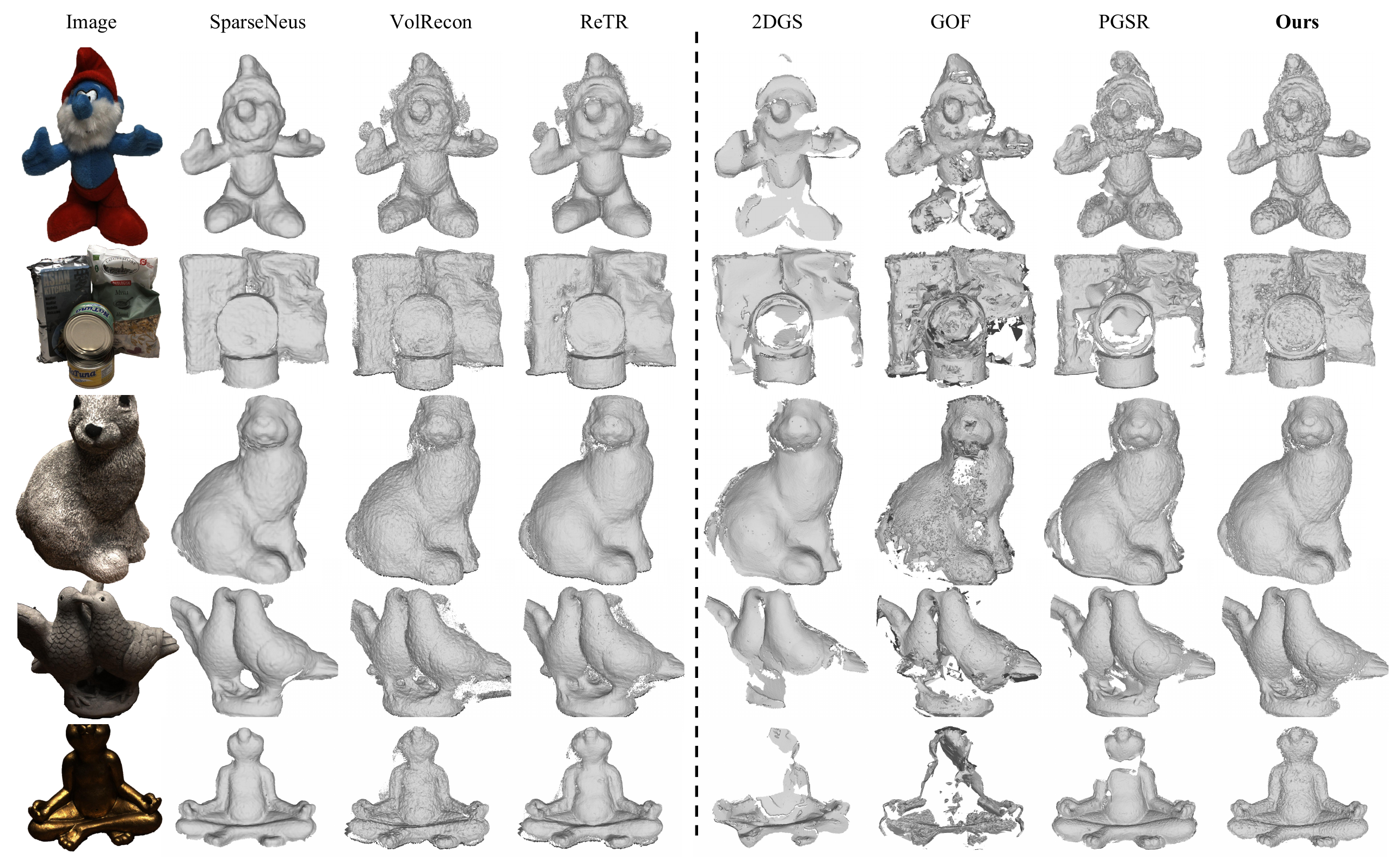}
\end{center}
\vspace{-15pt}
\caption{\textbf{DTU surface reconstruction results.} Our method achieves more complete reconstructions with finer details.}
\vspace{-13pt}
\label{fig:dtu_results}
\end{figure*}

\begin{figure}[t]
\begin{center}
\includegraphics[scale=0.56]{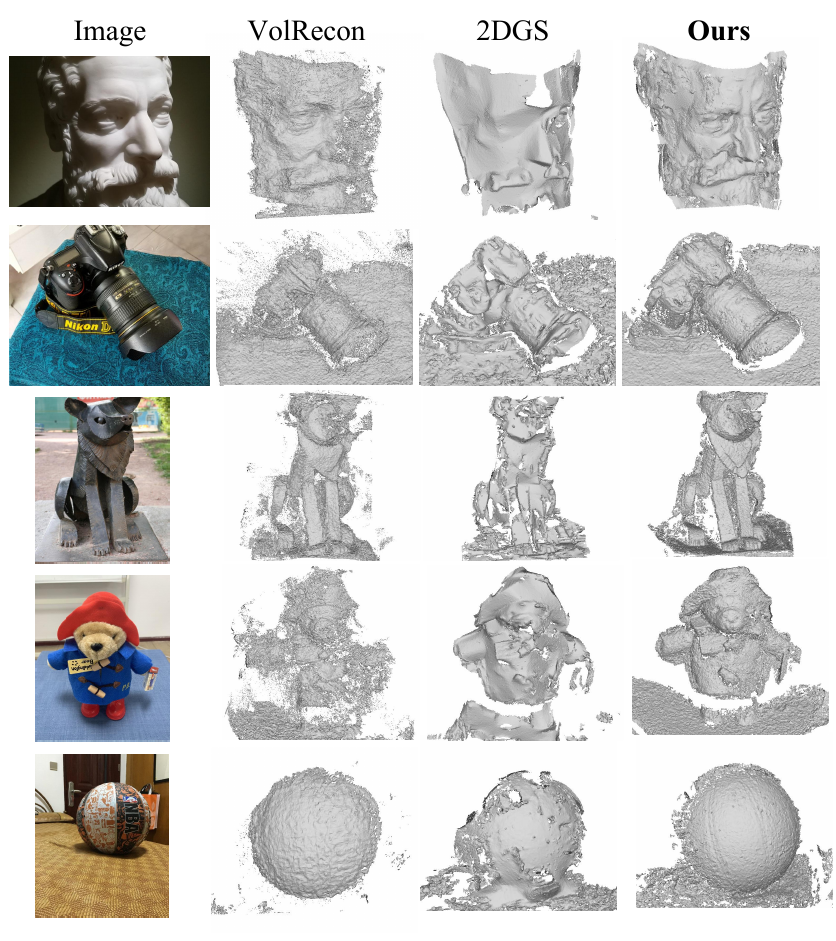}

\end{center}
\vspace{-15pt}
\caption{\textbf{BlendedMVS reconstruction results.} Our method achieves more complete and detailed reconstructions.}
\label{fig:bld_results}
\vspace{-10pt}
\end{figure}

\section{Experiments}
\label{sec:experiments}
In this section, we first introduce the implementation details and then provide quantitative and qualitative comparisons with current methods on DTU~\cite{aanaes2016large} dataset, as well as qualitative results on BlendedMVS~\cite{yao2020blendedmvs} dataset. We further analyze the contribution of each module and compare them with different methods through ablation studies.
\subsection{Datasets}
\label{sec:Datasets}
The DTU~\cite{aanaes2016large} dataset consists of 128 scenes captured in controlled lab environments using structured light scanners containing 49 or 64 images.
Similar to SparseNeuS~\cite{long2022sparseneus}, we select 15 scenes and use views 23, 24, and 33 for 3-view reconstruction, with an image resolution of 576×768.
We perform the quantitative evaluation on the DTU dataset using the official evaluation code to calculate the Chamfer Distance between the reconstructed mesh and the ground truth point cloud.
The BlendedMVS~\cite{yao2020blendedmvs}  dataset includes several large-scale real-world scenes.
We also show the qualitative reconstruction comparisons.
\subsection{Implementation Details}
\label{sec:implementation}
Our method is implemented based on 2DGS~\cite{huang20242d}.
For Gaussian primitive initialization, we use the point cloud back-projected depth maps from CLMVSNet~\cite{xiong2023cl}. Instead of using spherical harmonics to represent the color of the Gaussian primitives, we initialize the color using the RGB values from the image.
CLMVSNet uses an FPN for feature extraction, and we utilize the full-resolution feature maps with a channel dimension of 8 from the FPN last layer for feature splatting, 
The remaining parameters are initialized consistently with those in 2DGS.
We set the training iteration to 7000 and replaced the original 2DGS's adaptive density control with our proposed Gaussian update strategy. Similar to 2DGS~\cite{huang20242d}, we used truncated signed distance fusion (TSDF) to extract the final mesh.  
During TSDF fusion, the voxel size and truncated threshold are set to 0.004 and 0.02 respectively.
We set the hyperparameters of the loss function, $\lambda_1$, $\lambda_2$, $\lambda_3$, and $\lambda_4$, to 1000, 0.05, 1, and 0.2, respectively.

\begin{table}[t]
\renewcommand{\arraystretch}{1}
 \setlength{\tabcolsep}{3.7pt}
 \centering
 \scalebox{0.78}{
\begin{tabular}{cccccccccc}
\toprule
& 2DGS & MVS Init. & Fixed Color& Feat. Splat.& DGPR & SGU & CD $\downarrow$ \\ \hline
\textbf{(a)}& \usym{1F5F8}  &          &             &         &          &&2.813     \\
\textbf{(b)}& \usym{1F5F8}  &\usym{1F5F8}&             &         &         && 1.273     \\
\textbf{(c)}& \usym{1F5F8}  &\usym{1F5F8}&\usym{1F5F8}&         &    &   &1.212   \\
\textbf{(d)}& \usym{1F5F8}  &\usym{1F5F8}& &\usym{1F5F8}         &         &&1.224   \\
\textbf{(e)}& \usym{1F5F8} &\usym{1F5F8}&\usym{1F5F8}&\usym{1F5F8}&          & & 1.201   \\ \midrule
\textbf{(f)}& \usym{1F5F8} & \usym{1F5F8}&\usym{1F5F8}& \usym{1F5F8}        & \usym{1F5F8} &&   1.142    \\
\textbf{(g)}& \usym{1F5F8}  & \usym{1F5F8}&\usym{1F5F8}&\usym{1F5F8}&  & \usym{1F5F8}&     1.152 \\ \midrule
\textbf{(h)}& \usym{1F5F8}  &\usym{1F5F8}&\usym{1F5F8}&\usym{1F5F8}&\usym{1F5F8}  & \usym{1F5F8}     &\textbf{1.125}  \\ \bottomrule
\end{tabular}}

\caption{\textbf{Ablation on DTU~\cite{aanaes2016large} dataset.} DGPR and SGU represent proposed Direct Gaussian Primitive Regularization and Selective Gaussian Update respectively. Each module improves upon the baseline, leading to the best final results. 
}
\vspace{-10pt}
\label{table:abstudy}
\end{table}

\vspace{-5pt}
\subsection{Benchmark Comparisons}
We compare the sparse-view reconstruction performance of different NeRF-based and Gaussian-based methods on the DTU dataset. 
As shown in Tab. \ref{table:dtu_cd}, our method significantly improves upon Gaussian-based approaches under sparse views, outperforming baseline (2DGS) and previous Gaussian-based SOTA methods (PGSR) by 67\% and 46\%, respectively.
Our method also outperforms SparseNeus~\cite{long2022sparseneus}, the first sparse-view NeRF-based method with notable margins. Our method performs competitively with the state-of-the-art NeRF-based surface reconstruction methods.
The visualization results are shown in Fig. \ref{fig:dtu_results}.
Our method captures more surface details than implicit reconstruction approaches and outperforms Gaussian-based methods in preserving the complete scene structure. In comparison, Gaussian Splatting methods, \ie, 2DGS, GOF, and PGSR, tend to exhibit noise and missing areas.
Additionally, we performed a qualitative comparison on the BlendedMVS dataset. As shown in Fig. \ref{fig:bld_results}, our method achieves higher accuracy in reconstructing meshes for real-world scenes.

\begin{table}[t]
 \setlength{\tabcolsep}{2pt}
 \centering
  \scalebox{0.78}{
\begin{tabular}{lccccc}
\toprule
Methods   & 2DGS~\cite{huang20242d}  & PGSR~\cite{chen2024pgsr} & CLMVS.~\cite{xiong2023cl} & Ours \\ \midrule
3 view SfM Point Cloud     & 2.81 & 2.08 & -        & -    \\
49/64 views SfM Point Cloud & 1.65 & 1.53 & -        & -    \\
CLMVSNet Point Cloud          & 1.27 & 1.38 & 1.26     & \textbf{1.13} \\
\bottomrule
\end{tabular}}
\vspace{-3pt}
\caption{\textbf{Comparison of initialization strategies.} %
Initializing with MVS point cloud improves reconstruction performance. However, a plain combination (Chamfer Distance, CD:1.27) between 2DGS and CLMVSNet fails to surpass CLMVSNet results (CD: 1.26). Our method enhances both MVS and 2DGS through geometry-prioritized optimization.
}
\label{table:mvsvssfm}
\end{table}

\begin{figure}
\begin{center}
\includegraphics[scale=0.4]{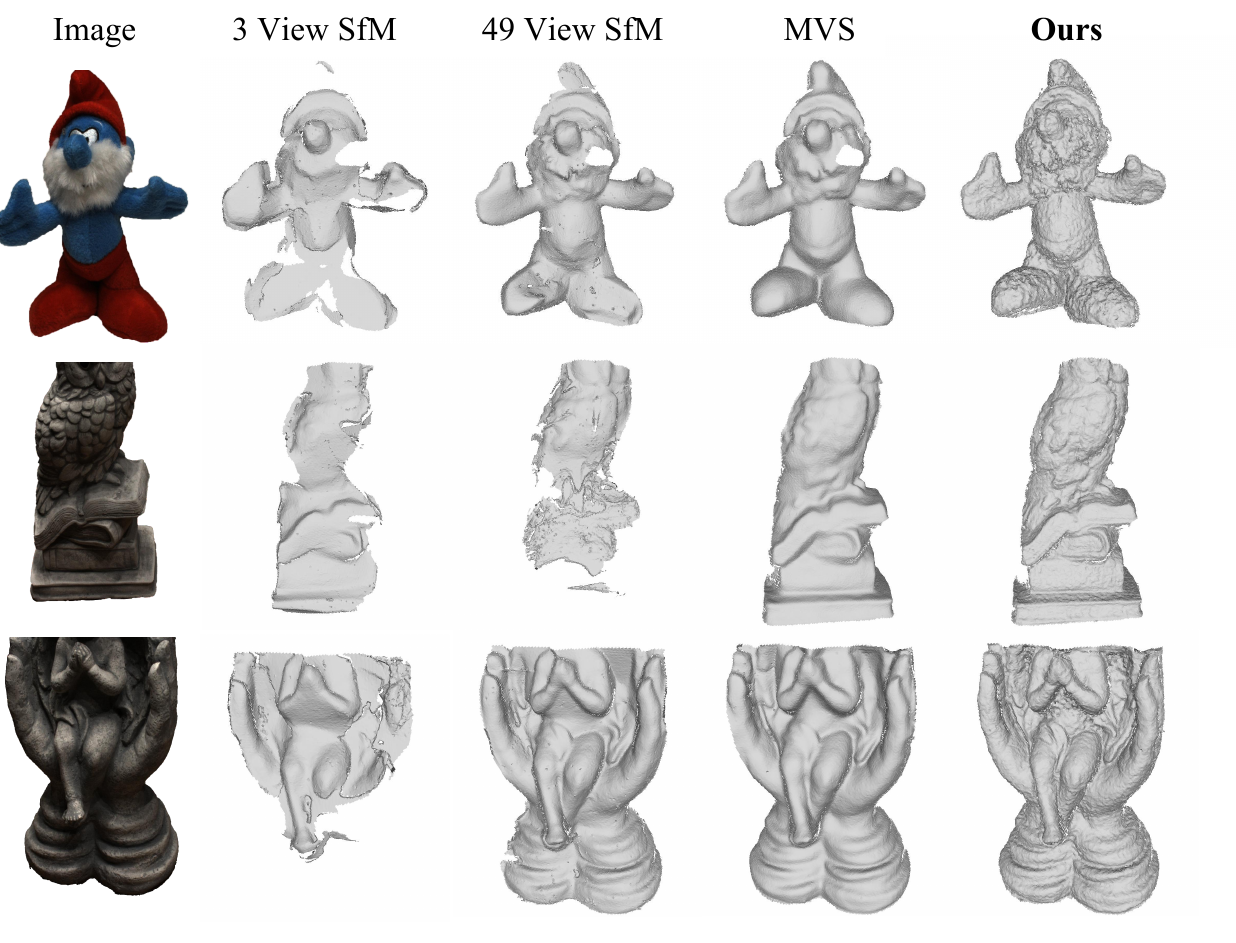}
\vspace{-30pt}
\end{center}
   \caption{\textbf{Comparison between different initialization strategies of 2DGS and our method.} %
   Initializing 2DGS with MVS point cloud improves SfM reconstruction completeness but struggles to capture details. Our method not only enhances completeness but also achieves detailed reconstruction.}
\label{fig:compare_init}
\vspace{-15pt}
\end{figure}

\subsection{Ablations and Analysis}
In this section, we perform ablation studies to analyze the effectiveness of different modules.
As shown in Tab. \ref{table:abstudy}, using MVS point cloud initialization notably enhances the performance (row \textbf{a}).
By fixing the colors and using feature splatting for geometric enhancement supervision, we further improve the accuracy and completeness of the reconstruction (row \textbf{e}).
The Direct Gaussian Primitive Regularization (DGPR) enhances the surface modeling capability of each Gaussian primitive (row \textbf{f}).
The Selective Gaussian Update (SGU) leverages rendering cues to enhance the precision of the Gaussian point set, further improving surface accuracy (row \textbf{g}).

\begin{table}[t]
\centering
\renewcommand{\arraystretch}{1}
 \setlength{\tabcolsep}{12pt}
\scalebox{0.78}{
\begin{tabular}{lccc}
\toprule
Settings    & Accuracy$\downarrow$ & Completion$\downarrow$ & Average$\downarrow$ \\ \midrule
SH=3        &     0.83     &   1.71         &      1.27   \\
SH=2        &     0.84     &  1.68          &      1.26   \\
SH=1        &    0.84      &  1.64          &     1.24    \\
Fixed Color &\textbf{0.81} &\textbf{1.61} &    \textbf{1.21}     \\ \bottomrule
\end{tabular}}
\vspace{-8pt}
\caption{\textbf{Ablation study on appearance learning.} Fixing the color helps in learning better geometry.
}
\vspace{-7pt}
\label{table:color}
\end{table}

\noindent{\textbf{Comparison of different initialization strategies.}} 
We compare the reconstruction results using various initialization strategies for Gaussian Splatting methods.
We show the results using 3 different initialization strategies: SfM point cloud from 3 viewpoints, SfM point cloud from 49 viewpoints, and the CLMVSNet reconstructed point cloud.
Using MVS point clouds achieves a more complete reconstruction than SfM.
Note that as shown in Tab. \ref{table:mvsvssfm}, due to the inherent ill-posedness in sparse-view reconstruction, the plain combination of MVS and 2DGS reconstruction yields suboptimal performance.
Even with CLMVSNet point clouds as initialization, the reconstruction quality still falls short of that achieved by CLMVSNet itself.
In contrast, our geometric-prioritized enhancement addresses this challenge, with results excelling both CLMVSNet and 2DGS with higher completeness and finer details (Fig.~\ref{fig:compare_init}).

\begin{figure}
\begin{center}
\includegraphics[scale=0.3]{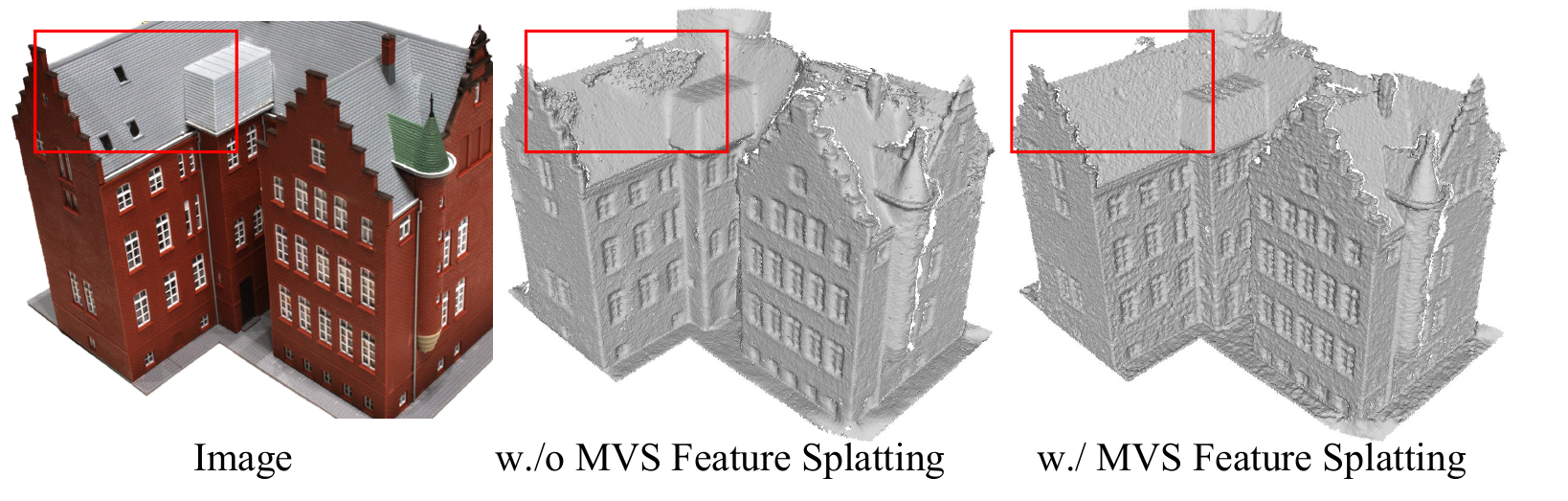}
\vspace{-30pt}
\end{center}
   \caption{\textbf{Visualization of MVS feature splatting.} Incorporating MVS feature splatting aids in achieving higher reconstruction accuracy in challenging regions. }
\label{fig:feature_splatting}
\vspace{-15pt}
\end{figure}

\noindent{\textbf{Geometrically enhanced supervision.}} 
In this paper, we mitigate appearance overfitting by fixing the color and feature parameters during optimization.
In the original Gaussian splatting, view-independent color is represented by spherical harmonics (SH) with learnable coefficients. Intuitively, reducing SH degrees can achieve a similar goal in alleviating color overfitting.
We compare the fixed color and reduced SH degrees in Tab. ~\ref{table:color}.
Results show that reducing the SH degrees also mitigates the degradation in geometry, with only 1 SH degree outperforming other settings (\ie, 2, 3) in surface reconstruction.
However, using learnable color coefficients still performs worse than the adopted choice of fixed color during training. 
Moreover, we enhance geometric supervision using MVS feature splatting. As shown in Fig. \ref{fig:feature_splatting}, MVS features are more expressive and provide more effective constraints, especially in challenging regions such as repetitive texture areas (\eg, the roof of the building).%

\begin{table}[t]
\renewcommand{\arraystretch}{1}
 \setlength{\tabcolsep}{6pt}
 \centering
 \scalebox{0.78}{
\begin{tabular}{lccc}
\toprule
Methods    & Accuracy$\downarrow$ & Compleness$\downarrow$ & Average$\downarrow$ \\ \midrule
Baseline        &0.790&1.612&    1.201         \\
Adaptive Density Control~\cite{kerbl20233d}        &0.814&1.654& 1.234          \\
Selective Gaussian Update        &\textbf{0.760}&\textbf{1.544}& \textbf{1.152}          \\
 \bottomrule
\end{tabular}}
\vspace{-8pt}
\caption{\textbf{Ablation on DTU~\cite{aanaes2016large} dataset.}
The proposed Selective Gaussian Update strategy enhances reconstruction accuracy and completeness compared to the generic Adaptive Density Control in sparse views.
}
\vspace{-7pt}
\label{table:gsupdate}
\end{table}

\noindent{\textbf{Ablation on Direct Gaussian Primitive Regularization.}} 
We analyze the impact of the reparameterization-based disk sampling number $K$ on the final results. As shown in Tab. \ref{table:sampling}, increasing the number of sampled points can slightly enhance the final reconstruction performance.

\begin{table}[t]
\centering
\renewcommand{\arraystretch}{1}
 \setlength{\tabcolsep}{17pt}
\scalebox{0.78}{
\vspace{-5pt}
\begin{tabular}{lccc}
\toprule
Settings    & Accuracy$\downarrow$ & Completion$\downarrow$ & Average$\downarrow$ \\ \midrule
$K$=9        &     0.746     &   1.544        &      1.145   \\
$K$=25        &     0.747   &  1.541        &      1.144  \\
$K$=49       &0.751 &1.532 &1.142    \\
$K$=81  &\textbf{0.747} &\textbf{1.531} &    \textbf{1.139}     \\ \bottomrule
\end{tabular}}
\vspace{-8pt}
\caption{\textbf{Comparison of different numbers of sampled points.} Increasing the sampling numbers lead to a slight performance improvement.
}
\vspace{-7pt}
\label{table:sampling}
\end{table}

\noindent\textbf{Selective Gaussian update strategy.} 
To validate the effectiveness of our selective update strategy, we compared it with the adaptive density control in the original 2DGS. 
Quantitative results are shown in the Tab. ~\ref{table:gsupdate}.
Due to the lack of sufficient multi-view supervision, the densified Gaussian primitives are prone to degradation, leading to declined reconstruction performance (CD:1.23) compared to the baseline (CD: 1.20).
The proposed selective Gaussian update strategy enhances the accuracy of the original Gaussian point set by utilizing high-precision points from the rendered cues, which further improves the surface reconstruction quality.

\begin{table}[t]
\renewcommand{\arraystretch}{1}
 \setlength{\tabcolsep}{8pt}
 \centering
 \scalebox{0.78}{
\begin{tabular}{lccccc}
\toprule
Methods & SparseNeus~\cite{long2022sparseneus} &NeuSurf~\cite{huang2024neusurf}             & Ours   \\ \midrule
Chamfer Distance & 1.27       &\textbf{0.99}               & 1.13   \\ \midrule
Training Time   & 25mins     &10 hours  & \textbf{10mins} \\ \bottomrule
\end{tabular}}

\vspace{-4pt}
\caption{\textbf{Performance and efficiency.} Our method achieves comparable performance while requiring the least training time.}
\label{table:performance_and_efficiency}
\vspace{-15pt}
\end{table}

\noindent{\textbf{Performance and efficiency.}} 
We compared reconstruction performance and training time with previous state-of-the-art surface reconstruction methods. As shown in Tab. \ref{table:performance_and_efficiency}, our method achieves better performance and $2\times$ faster training speed than SparseNeus~\cite{long2022sparseneus}.
Meanwhile, our method achieves competitive performance while being $60\times$ faster than NeuSurf~\cite{huang2024neusurf}.

\section{Conclusion}
In this paper, we propose Sparse2DGS, addressing the challenge of ill-posed geometry optimization from sparse views through geometry-prioritized enhancement schemes.
Our method not only reconstructs accurate and complete scene surfaces but also achieves higher efficiency than NeRF-based methods.
\textit{Limitation}: Despite their effectiveness in dense matching, MVS methods inevitably encounter degradation under occlusions. The error in occluded areas can propagate to our method, leading to potential performance declines.
\par
{\noindent \textbf{Acknowledgements}
Y. Zhang was supported by NSFC (No.U19B2037) and the Natural Science Basic Research Program of Shaanxi (No.2021JCW-03). Y. Zhu was supported by NSFC (No.61901384).}

\vspace{-15pt}
{
    \small
    \bibliographystyle{ieeenat_fullname}
    \bibliography{main}
}

\end{document}